%% file: arxiv.tex
\documentclass[letterpaper]{article} % DO NOT CHANGE THIS
\usepackage[preprint]{aaai2027}  % arXiv version: show authors, no AAAI copyright notice
\usepackage[hyphens]{url}  % DO NOT CHANGE THIS
\usepackage{graphicx} % DO NOT CHANGE THIS
\usepackage[numbers,sort&compress]{natbib}  % arXiv version: CVPR-style numeric citations
\usepackage{caption} % DO NOT CHANGE THIS AND DO NOT ADD ANY OPTIONS TO IT
\usepackage{algorithm}
\usepackage{algorithmic}

\usepackage{newfloat}
\usepackage{listings}
\DeclareCaptionStyle{ruled}{labelfont=normalfont,labelsep=colon,strut=off} % DO NOT CHANGE THIS
\floatstyle{ruled}
\newfloat{listing}{tb}{lst}{}
\floatname{listing}{Listing}

\usepackage{booktabs}

\usepackage{xspace}
\usepackage{booktabs}
\usepackage{multirow}
\usepackage{amsmath}
\usepackage{amssymb}
\usepackage{subcaption}

\IfFileExists{arxiv/arxiv.tex}{\def\arxivdir{arxiv/}}{\def\arxivdir{}}
\graphicspath{{\arxivdir}}

\usepackage[pagebackref,breaklinks,colorlinks,allcolors=blue]{hyperref}

\newcommand{\method}{\textsc{SEAL}\xspace}

\newcommand{\codeurl}{\url{https://github.com/Lvxan/SEAL}}

\title{Location Matters: Sparse Latent-Variance Attacks for Face Privacy against Diffusion Personalization}
\author{
    Shuting Yang\textsuperscript{\rm 1},
    Shuxin Li\textsuperscript{\rm 2},
    Linyan Dai\textsuperscript{\rm 3},
    Xinwei Zhang\textsuperscript{\rm 3},
    Dong Huang\textsuperscript{\rm 1},\\
    Renyang Liu\textsuperscript{\rm 1,}\corresponding,
    See-Kiong Ng\textsuperscript{\rm 1}
}
\affiliations{
    \textsuperscript{\rm 1}National University of Singapore\\
    \textsuperscript{\rm 2}Nanyang Technological University\\
    \textsuperscript{\rm 3}The Hong Kong Polytechnic University
}
\correspondingemail{ryliu@nus.edu.sg}  % shown in the first-page footnote

\makeatletter\providecommand\@LN[2]{}\makeatother

\begin{document}

\maketitle
\thispagestyle{plain} 

\input{sections/00_abstract}
\input{sections/01_Intro}

\input{sections/02_Related_Work}
\input{sections/04_Method}
\input{sections/05_Experiments}

\input{sections/06_Conclusion}

\bibliographystyle{plainnat}
\bibliography{\arxivdir aaai2027}

\clearpage
\appendix
\pagestyle{plain}  % page numbers in the appendix

\input{sections/08_Appendix}

\end{document}

%% file: sections/00_abstract.tex
\begin{abstract}
Text-to-image diffusion models can learn and reproduce an individual's identity from only a few reference photos, creating risks of unauthorized and misleading content generation.
Proactive defenses address this risk by adding adversarial perturbations to facial images before release, thereby disrupting identity acquisition during downstream personalization.
However, existing methods typically rely on dense $L_\infty$-bounded perturbations, whose accumulated distortions can become visible around facial structures. 
Recent methods improve protected-image quality, but often at the cost of reduced protection effectiveness.
In this paper, we propose \textbf{S}parse \textbf{E}ncoder \textbf{A}ttack on \textbf{L}atent Variance (\method), a sparse adversarial protection framework that modifies only 5\% of the spatial locations in a facial image.
\method combines a differentiable top-$k$ masking scheme that jointly optimizes perturbation locations and values with a variance-oriented objective over the frozen VAE encoder's latent distribution to disrupt identity-consistent representations.
A saliency-guided dispersion penalty further discourages perturbation concentration in perceptually salient facial regions.
Experiments on CelebA-HQ and VGGFace2 show that \method achieves the best protected-image quality among the compared methods, exceeding 41 dB PSNR with 0.97 SSIM and at most 0.07 LPIPS, while reducing identity similarity to at most 0.15 after diffusion personalization.
These results demonstrate that sparse location optimization enables strong identity protection while preserving high visual quality.
Code and data are available at \codeurl.
\end{abstract}

%% file: sections/01_Intro.tex
\section{Introduction}
Text-to-image diffusion models~\cite{ho2020ddpm,rombach2022ldm} have demonstrated remarkable capabilities in generating photorealistic images from natural language descriptions.
Personalization techniques, such as DreamBooth~\cite{ruiz2023dreambooth} and Textual Inversion~\cite{gal2023textualinversion}, further enable these models to reproduce a specific individual from only a few reference photos.
While this capability facilitates customized content creation, it also lowers the barrier to unauthorized identity replication.
An adversary can collect publicly available facial images, personalize a diffusion model to the depicted individual, and generate misleading or harmful content involving that person~\cite{qu2023unsafe}.
This risk motivates proactive protection mechanisms that prevent released facial images from being used to learn and reproduce an individual's identity.

\begin{figure}[t]
  \centering
  \includegraphics[width=\linewidth]{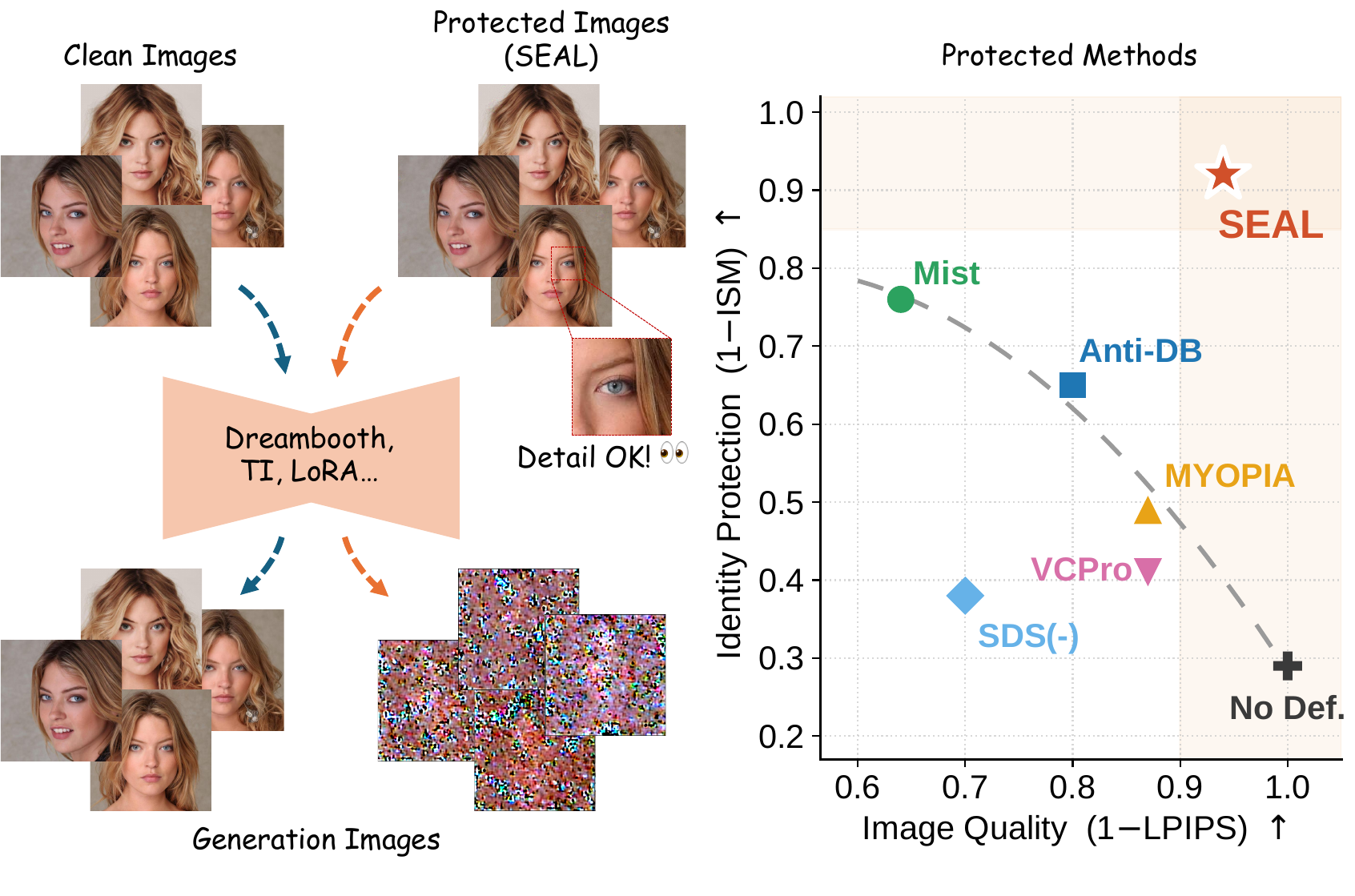}
  \caption{By modifying only a small fraction of pixels, \method preserves the visual appearance of the protected face image while disrupting identity
  reproduction after diffusion personalization. Compared with existing protection methods, \method achieves both high image quality and strong protection effectiveness.}
  \label{fig:teaser}
\end{figure}

A common proactive strategy is to add bounded adversarial perturbations to facial images before their release, such that personalized models fine-tuned on these protected images fail to reproduce the target identity.
Early approaches construct such perturbations by maximizing the diffusion denoising loss, as in AdvDM~\cite{liang2023advdm}, or by shifting image representations toward a predefined latent target, as in Mist~\cite{liang2023mist}.
Anti-DreamBooth~\cite{le2023antidreambooth} instead directly maximizes the DreamBooth training loss to interfere with personalization. 
More recent methods, including PID~\cite{li2024pid} and Variance as a Catalyst~\cite{yang2025variance}, manipulate the mean or variance of the VAE encoder's latent distribution, revealing that disrupting latent statistics can effectively impair identity acquisition across different prompts.
Collectively, these studies establish adversarial perturbation as a viable defense against unauthorized diffusion personalization.

However, most existing methods apply dense perturbations across every pixel, and the accumulated distortions can become perceptible around visually sensitive facial structures.
Recent works have sought to improve perceptual quality: VCPro~\cite{mi2026vcpro} uses concept masks to guide a regional adversarial objective, whereas MYOPIA~\cite{wu2025myopia} exploits local information to conceal perturbation artifacts. However, these improvements in visual quality are accompanied by reduced protection effectiveness.
Overall, existing methods exhibit a pronounced trade-off between protection and visual quality. As shown in Figure~\ref{fig:teaser}, methods that achieve stronger protection often rely on dense perturbations that noticeably degrade image quality, whereas visual-friendly methods preserve quality at the cost of reduced protection effectiveness. Achieving strong protection while maintaining high visual quality therefore remains an open challenge.

To address this challenge, we propose \textbf{S}parse \textbf{E}ncoder \textbf{A}ttack on \textbf{L}atent Variance (\method), a sparse perturbation framework for proactive face privacy protection.
Instead of distributing perturbations densely across the entire image, \method jointly optimizes a bounded perturbation $\delta$ and a binary mask $m$ that determines the perturbed pixel locations.
To handle the discrete pixel-selection process, we adapt the differentiable top-$k$ masking mechanism of Sparse-PGD~\cite{zhong2024sparsepgd}, which jointly optimizes perturbation values and spatial locations under a fixed support budget.
Within the selected sparse support, \method maximizes a Laplace objective over the variance of the VAE encoder distribution, following prior evidence that disrupting latent statistics can interfere with the acquisition of subject-specific identity information~\cite{li2024pid,yang2025variance}.
However, sparse optimization may concentrate perturbations around salient facial structures, making local artifacts noticeable even when the overall perturbation support is small.
We therefore introduce a saliency-guided dispersion regularizer that penalizes high local perturbation density in perceptually salient facial regions.
Together, these components allow \method to allocate a limited perturbation budget to locations that effectively disrupt personalization while preserving the visual quality of the protected image.

We evaluate \method on CelebA-HQ~\cite{karras2018celeba} and VGGFace2~\cite{cao2018vggface2} using Stable Diffusion v2.1 personalized with DreamBooth under two primary inference prompts.
Quantitatively, with only 5\% of pixels modified, \method achieves LPIPS $\leq$ $0.07$ on both datasets, while keeping the identity similarity at most $0.15$ under both prompts and raising the face detection failure rate to $0.93$ under \textit{a photo of sks person}.
This improvement is mainly due to the way the locations of the perturbations are placed. 
Ablation experiments show that when the 5\% locations are chosen at random rather than optimized, ISM rises from 0.07 to 0.68.
Qualitatively, as shown in Figure~\ref{fig:qualitative}, the protected images preserve high visual quality in facial-detail regions, while DreamBooth generations under different prompts are severely disrupted.
Our main contributions are as follows:

\begin{itemize}
\item We propose \method, a sparse perturbation approach to proactive face identity protection from diffusion personalization that modifies only a small fraction of the pixels in each protected image.

\item We develop a joint optimization framework that combines differentiable top-$k$ pixel selection with a latent-variance disruption objective, enabling perturbation values and locations to be optimized simultaneously under an $L_0$ support budget and an $L_\infty$ magnitude constraint.

\item We introduce a saliency-guided dispersion regularizer that discourages perturbation clustering in perceptually salient facial regions while retaining the protection capacity of the sparse perturbation.

\item Extensive experiments on CelebA-HQ and VGGFace2 demonstrate that \method achieves the best protected-image quality and the lowest identity similarity among the compared methods, while maintaining competitive computational cost.

\end{itemize}

%% file: sections/02_Related_Work.tex
\section{Related Work}

\subsection{Personalized Text-to-Image Generation}

Text-to-image diffusion models have become a dominant paradigm for image synthesis, with latent diffusion models~\cite{rombach2022ldm} improving computational efficiency by performing denoising in a compressed latent space. Beyond generic generation, personalization techniques adapt pretrained diffusion models to reproduce a specific subject or visual concept from only a few reference images. Textual Inversion~\cite{gal2023textualinversion} learns a new token embedding to represent the target visual concept, whereas DreamBooth~\cite{ruiz2023dreambooth} fine-tunes the generative model to bind a unique identifier to the target subject. Custom Diffusion~\cite{kumari2023customdiffusion} enables multi-concept customization, and LoRA~\cite{hu2022lora} reduces fine-tuning cost.
When applied to human faces, however, such personalization capabilities raise serious privacy and security concerns, as publicly available photographs can be exploited to generate misleading or unauthorized depictions of specific individuals~\cite{qu2023unsafe}.

\begin{figure*}[t]
\centering
\includegraphics[width=\textwidth]{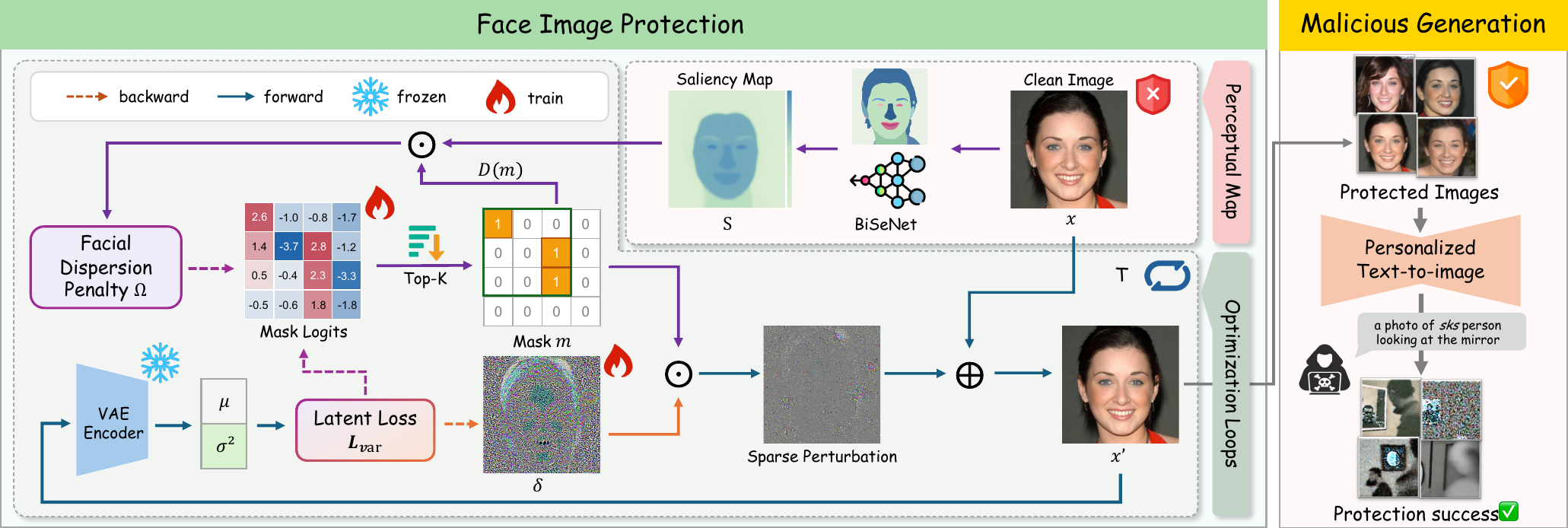}
\caption{
Overview of \method.
Given a clean image $x$, \method jointly optimizes a sparse binary mask $m$ and a bounded perturbation $\delta$ to disrupt the VAE latent variance.
A saliency-guided dispersion regularizer discourages the selected pixels from clustering in perceptually salient facial regions.
}
\label{fig:framework}
\end{figure*}

\subsection{Image Protection Against Personalization}
To mitigate unauthorized personalization, most protection methods add adversarial perturbations to images before release to disrupt their use in downstream diffusion personalization, though a separate line instead pursues privacy on the model side by training identity-decoupled personalized models~\cite{dai2026iddm}.
These input-side methods differ primarily in the model component or training objective targeted during perturbation optimization.
AdvDM~\cite{liang2023advdm} maximizes the diffusion denoising loss, while PhotoGuard~\cite{salman2023photoguard} introduces an encoder attack that shifts the latent representation toward a predefined target.
Mist~\cite{liang2023mist} combines denoising and latent-target objectives to disrupt both the image representation and the denoising process.
Other methods manipulate the latent distribution produced by the VAE encoder: PID~\cite{li2024pid} jointly perturbs its mean and log-variance, whereas Variance as a Catalyst~\cite{yang2025variance} develops distribution-oriented objectives, including a Laplace-based loss, to amplify latent variance and disrupt identity acquisition.
Personalization-aware methods instead incorporate the downstream training process into perturbation optimization.
Anti-DreamBooth~\cite{le2023antidreambooth} simulates DreamBooth fine-tuning and maximizes its training loss, while MetaCloak~\cite{liu2024metacloak} employs meta-learning and transformation-aware optimization to improve perturbation transferability and robustness.
Several recent works further focus on the visual quality of protected images.
SDS(-)~\cite{xue2024sds} minimizes the semantic loss to produce more natural-looking perturbations.
MYOPIA~\cite{wu2025myopia} combines just-noticeable-difference constraints with expectation over transformations to regulate perturbation location and intensity while improving robustness.
Latent Diffusion Unlearning~\cite{devulapally2025ldu} constructs
trajectory-shifted perturbations in latent space by alternating denoising and inversion, thereby improving the visual quality of protected images.
VCPro~\cite{mi2026vcpro} employs selective adversarial perturbations and a relaxed optimization objective to improve the trade-off between perturbation visibility and concept protection.
Despite these advances, these methods treat perturbation placement as a visual or region-level constraint, leaving pixel-level support selection outside the optimization objective that governs protection effectiveness.
Sparse $L_0$-constrained attacks such as Sparse-PGD~\cite{zhong2024sparsepgd} show that adversarial effects can be achieved by modifying only a small subset of pixels.
Building on this sparse optimization principle, \method combines differentiable pixel selection with latent-variance disruption and saliency-guided dispersion regularization for proactive face identity protection.

%% file: sections/04_Method.tex
\section{Method}

\subsection{Problem Formulation}
\label{sec:problem}

Given a facial image $x \in [0,1]^{3 \times H \times W}$, the defender releases a protected image $x'$ obtained by adding a perturbation $\delta \in \mathbb{R}^{3 \times H \times W}$ bounded by $\|\delta\|_\infty \leq \epsilon$.
Protection succeeds when $x'$ stays visually indistinguishable from $x$ while a model personalized on $x'$ fails to reproduce the identity.
As the surrogate we use the frozen VAE encoder $\mathcal{E}_\theta$ of a latent diffusion model~\cite{rombach2022ldm}, which maps an image to a diagonal Gaussian with mean $\mu(\cdot)$ and variance $\sigma^2(\cdot)$.

\paragraph{Attacker.}
The attacker collects publicly available facial images of a target individual and fine-tunes a diffusion model on them with a personalization algorithm such as DreamBooth~\cite{ruiz2023dreambooth}, whose customized prompt binds a unique identifier such as ``sks'' to the subject.
The personalized model is then queried to render that individual under arbitrary prompts and contexts.

\paragraph{Defender.}
The defender may modify images only before release and has no control over their subsequent collection or use.
The defender also has no access to the attacker's model, personalization algorithm, training hyperparameters, identifier token, or inference prompts, and must therefore construct the protection on an accessible surrogate model.

\subsection{Overview}

\method inherits the setting of Sec.~\ref{sec:problem} and adds a restriction: rather than letting $\delta$ act on every pixel, it confines the perturbation to $k$ spatial locations selected by a binary mask $m \in \{0,1\}^{H \times W}$, shared across the three color channels.
The released image is therefore $x'=\operatorname{clip}(x+\delta\odot m,0,1)$.
We further construct a perceptual saliency map $S(x) \in [0,1]^{H \times W}$, in which larger values indicate facial regions where perturbations are more visually conspicuous.
The conceptual optimization objective is:
\begin{equation}
\begin{aligned}
\max_{\delta,\,m}\quad
&\mathcal{L}_{\mathrm{var}}(x')-\lambda\,\Omega\!\left(m,S(x)\right),\\
\mathrm{s.t.}\quad
&\|\delta\|_\infty\leq\epsilon,\qquad
\|m\|_0=k.
\end{aligned}
\label{eq:objective}
\end{equation}
Here, $\mathcal{L}_\mathrm{var}$ disrupts the latent distribution of the protected image, while $\Omega$ discourages perturbation concentration in perceptually salient facial regions.
Visual fidelity is controlled jointly by the sparse support, the perturbation-magnitude bound, and the saliency-guided dispersion term.
Eq.~\eqref{eq:objective} captures the desired trade-off, while the normalized mask-update rule used in optimization is specified in Sec.~\ref{sec:dispersion}.

As illustrated in Figure~\ref{fig:framework}, \method consists of three components.
First, we adopt a Laplace-based objective over the VAE latent variance to disrupt the identity acquisition during personalization (Sec.~\ref{sec:laplace}).
% as the protection objective $\mathcal{L}_\mathrm{var}$ (Sec.~\ref{sec:laplace}).
Second, we adapt a differentiable top-$k$ masking mechanism to jointly optimize the perturbation values and their spatial support under an $L_0$ budget (Sec.~\ref{sec:sparse}).
Third, we introduce a saliency-guided dispersion regularizer that discourages sparse perturbations from concentrating in perceptually salient facial regions (Sec.~\ref{sec:dispersion}).

\subsection{Latent Variance Disruption}
\label{sec:laplace}

The frozen VAE encoder $\mathcal{E}$ maps a protected image $x'$ to a diagonal Gaussian posterior parameterized by its mean $\mu(x')$ and variance $\sigma^2(x')$.
During diffusion personalization, the encoded latent provides the image-side training signal from which the model learns subject-specific information.
Prior work has shown that manipulating the latent statistics of protected images can impair identity acquisition during personalization~\cite{li2024pid,yang2025variance}.
We therefore maximize the latent variance of $x'$ to increase uncertainty in the encoded representation and disrupt the identity-consistent latent signal used for personalization.

Specifically, we adopt the Laplace-based objective from Variance as a Catalyst~\cite{yang2025variance}:
\begin{equation}
\mathcal{L}_\mathrm{var}(x') = \sum_{j} \frac{\left|\sigma_j^2(x') - \mu_0\right|}{b},
\label{eq:laplace}
\end{equation}
where $j$ indexes all spatial and channel dimensions of the latent variance, and $\mu_0$ and $b$ are the location and scale parameters, respectively.
Following~\cite{yang2025variance}, we set $\mu_0=0$ and $b=1$.
Since $\sigma_j^2(x')\geq0$, Eq.~\eqref{eq:laplace} is equivalent to maximizing the total latent variance.
Its derivative with respect to $\sigma_j^2$ is $1/b$ almost everywhere, providing a non-saturating optimization signal across latent dimensions.
When the encoder parameterizes the posterior using log-variance, we recover the variance as $\sigma^2=\exp(\log\sigma^2)$ before evaluating Eq.~\eqref{eq:laplace}.

\subsection{Differentiable Sparse Masking}
\label{sec:sparse}

The cardinality constraint $\|m\|_0 = k$ is discrete and non-differentiable, preventing direct gradient-based optimization of the binary mask.
We address this by introducing continuous mask logits $\ell \in \mathbb{R}^{H \times W}$ and adapting the differentiable top-$k$ masking mechanism of Sparse-PGD~\cite{zhong2024sparsepgd}.
In the forward pass, we compute element-wise soft scores $\hat{m}_i = \bigl(1 + e^{-\ell_i}\bigr)^{-1}$ and project them to a binary mask by retaining the top-$k$ entries:
\begin{equation}
m = \operatorname{TopK}(\hat{m},\, k),
\label{eq:topk}
\end{equation}
where $\operatorname{TopK}(\hat{m}, k)$ returns a binary mask whose $k$ largest entries are one and whose remaining entries are zero.
The resulting hard mask determines the spatial support of the perturbation $\delta \odot m$.

Since the top-$k$ projection is non-differentiable, we employ a straight-through estimator (STE).
The forward pass uses the hard mask $m$, whereas the backward pass approximates its gradient using the soft mask $\hat{m}$.
Because one spatial mask value is shared across the three color channels, the approximate gradient of $\mathcal{L}_{\mathrm{var}}$ with respect to $\ell_i$ is

\begin{equation}
\frac{\partial\mathcal{L}_{\mathrm{var}}}{\partial\ell_i}
\approx
\frac{1}{3}
\left(
\sum_{c=1}^{3}
\frac{\partial\mathcal{L}_{\mathrm{var}}}
{\partial(\delta_{c,i}m_i)}
\delta_{c,i}
\right)
\hat{m}_i(1-\hat{m}_i).
\label{eq:ste}
\end{equation}

At each iteration, the perturbation $\delta$ is updated using a projected gradient-ascent step~\cite{madry2018pgd}:
\begin{equation}
\delta \leftarrow \Pi_{[-\epsilon,\epsilon]}\left(\delta + \alpha \cdot \mathrm{sign}\!\left(\nabla_\delta \mathcal{L}_\mathrm{var}\right)\right),
\label{eq:pgd}
\end{equation}
where $\alpha$ is the perturbation step size and $\Pi_{[-\epsilon,\epsilon]}$
denotes the element-wise projection onto the $L_\infty$ constraint.
The mask logits are updated via gradient ascent:
% with step size $\beta$:
\begin{equation}
\ell \leftarrow \ell + \sqrt{HW}\,\beta \cdot g_\ell,
\label{eq:mask_update}
\end{equation}
where $\beta$ is the mask step size and $g_\ell$ is defined in Sec.~\ref{sec:dispersion}.

To prevent the optimization from becoming trapped when the mask stabilizes prematurely, we employ patience-based reinitialization.
If the hard mask $m$ remains unchanged for $P$ consecutive iterations, both $\ell$ and $\delta$ are randomly reinitialized.
We keep the perturbation with the highest objective $\mathcal{L}_\mathrm{var}-\lambda\Omega$ across all iterations as the released protection.

\begin{figure*}[t]
\centering
\includegraphics[width=\textwidth]{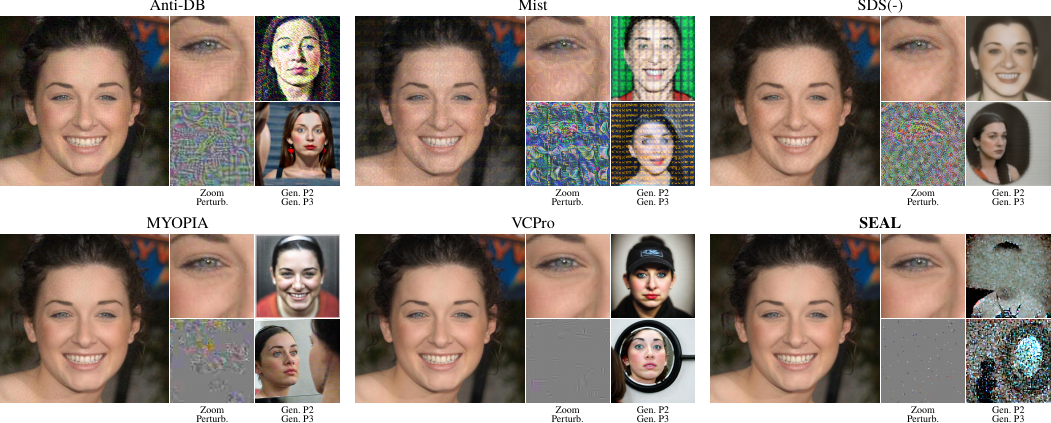}
\caption{
Qualitative comparison.
Each method shows the protected image, a magnified crop with its perturbation pattern, and DreamBooth generations under two prompts.
P2: \textit{a dslr portrait of sks person}; P3: \textit{a photo of sks person looking at the mirror}.
}
\label{fig:qualitative}
\end{figure*}

\subsection{Saliency-Guided Dispersion}
\label{sec:dispersion}

The sparse mask learned in Sec.~\ref{sec:sparse} tends to concentrate perturbations in regions that yield strong latent disruption. 
For facial images, these regions may coincide with perceptually sensitive structures, such as the eyes, nose, and mouth, making local perturbation artifacts more conspicuous.
To counteract this effect, we introduce a saliency-guided dispersion penalty that discourages high local perturbation density in salient facial regions.

\paragraph{Saliency map.}
We parse $x$ using a BiSeNet-based face parser~\cite{yu2018bisenet} with the 19-class CelebAMask-HQ taxonomy~\cite{lee2020maskgan}.
Studies of face viewing show that gaze is primarily allocated to internal facial features, particularly the eyes, nose, and mouth, with the eyes playing a preferential role in guiding fixations and face identification~\cite{hessels2020gaze,han2021peripheral}.
Based on these findings, we construct the saliency map $S(x)$ by assigning each semantic class a fixed weight $w_c \in [0,1]$, where larger values impose stronger dispersion penalties on perceptually salient regions: $1.0$ for the eyes, $0.9$ for the nose and mouth, $0.8$ for the skin, $0.02$ for the hair, and $0.0$ for the background.
The weighted label map is Gaussian-smoothed and min-max normalized to $[0,1]$ to reduce boundary artifacts, with the complete class-to-weight mapping reported in Table~\ref{tab:saliency_weights} in the Appendix.

\paragraph{Dispersion penalty.}
We measure the local density of the mask $m$ using a normalized $K \times K$ mean-pooling kernel $\mathbf{W}$, whose entries equal $1/K^2$:
\begin{equation}
D(m) = m * \mathbf{W},
\label{eq:density}
\end{equation}
where $*$ denotes convolution with padding that preserves the spatial resolution.
Thus, $D(m)_i\in[0,1]$ represents the fraction of selected pixels within the local neighborhood centered at location $i$.

The dispersion penalty is defined as:
\begin{equation}
\Omega(m, S) = \frac{1}{HW} \sum_{i} \mathrm{ReLU}\!\left(D(m)_i \cdot S_i - \tau\right),
\label{eq:dispersion_penalty}
\end{equation}
where $\tau$ is a density threshold.
The penalty remains zero when the saliency-weighted local density does not exceed $\tau$ and increases with violations above the threshold.

\paragraph{Gradient-balanced optimization.}
Since $\Omega$ depends only on the mask, the perturbation $\delta$ is updated using $\mathcal{L}_{\mathrm{var}}$ as in Eq.~\eqref{eq:pgd}.
For the mask logits, directly combining the two raw gradients can cause the update to be dominated by the term with the larger gradient magnitude.
We therefore normalize the two gradient components independently before combining them:
\begin{equation}
g_\ell
=
\frac{\nabla_\ell\mathcal{L}_{\mathrm{var}}}
{\|\nabla_\ell\mathcal{L}_{\mathrm{var}}\|_2+\eta}
-
\lambda
\frac{\nabla_\ell\Omega}
{\|\nabla_\ell\Omega\|_2+\eta},
\label{eq:grad_balance}
\end{equation}
where $\eta>0$ is a small constant for numerical stability.
For $\Omega$, the gradient is propagated through the hard mask with a straight-through estimator that treats the mask as an identity map of $\ell$, bypassing the sigmoid so that the dispersion gradient does not vanish once the logits saturate.
After normalization, $\lambda$ controls the relative contribution of latent disruption and saliency-guided dispersion to the mask update.

The complete optimization procedure is provided in Algorithm~\ref{alg:ours} in Appendix~\ref{app:algorithm}.

%% file: sections/05_Experiments.tex
\section{Experiments}

\begin{table*}[t]
\centering
\small
\setlength{\tabcolsep}{2pt}
\begin{tabular}{ll cccc cccc cccc}
\toprule
 & & \multicolumn{4}{c}{Image Quality}
 & \multicolumn{4}{c}{Protection (P1)}
 & \multicolumn{4}{c}{Protection (P2)} \\
\cmidrule(lr){3-6} \cmidrule(lr){7-10} \cmidrule(lr){11-14}
Dataset & Method
 & SSIM$\uparrow$ & PSNR$\uparrow$ & LPIPS$\downarrow$ & DISTS$\downarrow$
 & BRISQUE$\uparrow$ & SER-FIQ$\downarrow$ & FDFR$\uparrow$ & ISM$\downarrow$
 & BRISQUE$\uparrow$ & SER-FIQ$\downarrow$ & FDFR$\uparrow$ & ISM$\downarrow$ \\
\midrule
\multirow{7}{*}{VGGFace2}
 & No Def.    & -- & -- & -- & -- & 17.34 & 0.75 & 0.01 & 0.67 & 8.60 & 0.72 & 0.01 & 0.49 \\
 & Anti-DB       & 0.90 & 34.52 & 0.21 & 0.18 & 35.26 & \underline{0.35} & \underline{0.32} & 0.35 & 40.28 & \textbf{0.40} & \textbf{0.43} & 0.27 \\
 & Mist          & 0.74 & 28.62 & 0.38 & 0.38 & 20.59 & 0.47 & 0.12 & \underline{0.22} & 32.84 & \underline{0.44} & 0.23 & \underline{0.19} \\
 & SDS(-)        & 0.74 & 29.87 & 0.31 & 0.34 & \underline{40.84} & 0.75 & 0.00 & 0.62 & \underline{47.16} & 0.77 & 0.00 & 0.54 \\
 & MYOPIA        & \underline{0.92} & 34.35 & \underline{0.11} & 0.22 & 31.25 & 0.72 & 0.00 & 0.55 & \textbf{54.28} & 0.77 & 0.02 & 0.51 \\
 & VCPro         & 0.90 & \underline{36.40} & 0.16 & \underline{0.17} & 21.47 & 0.71 & 0.01 & 0.59 & 18.02 & 0.73 & 0.02 & 0.44 \\
 & \textbf{\method} & \textbf{0.97} & \textbf{41.08} & \textbf{0.07} & \textbf{0.07} & \textbf{76.89} & \textbf{0.06} & \textbf{0.93} & \textbf{0.11} & 34.43 & 0.50 & \underline{0.33} & \textbf{0.15} \\
\midrule
\multirow{7}{*}{CelebA-HQ}
 & No Def.    & -- & -- & -- & -- & 19.78 & 0.75 & 0.00 & 0.71 & 8.38 & 0.73 & 0.02 & 0.47 \\
 & Anti-DB       & 0.90 & 34.34 & 0.20 & 0.19 & 35.46 & \underline{0.31} & \underline{0.40} & 0.35 & 39.83 & \textbf{0.40} & \textbf{0.45} & 0.24 \\
 & Mist          & 0.75 & 28.56 & 0.36 & 0.39 & 17.75 & 0.56 & 0.05 & \underline{0.24} & 32.93 & 0.49 & 0.10 & \underline{0.16} \\
 & SDS(-)        & 0.74 & 29.81 & 0.30 & 0.35 & \underline{37.70} & 0.76 & 0.00 & 0.62 & \underline{41.31} & 0.77 & 0.00 & 0.52 \\
 & MYOPIA        & 0.91 & 33.83 & \underline{0.13} & 0.24 & 31.98 & 0.70 & 0.00 & 0.51 & \textbf{57.68} & 0.77 & 0.02 & 0.49 \\
 & VCPro         & \underline{0.92} & \underline{37.40} & \underline{0.13} & \underline{0.17} & 34.72 & 0.72 & 0.01 & 0.59 & 18.28 & 0.73 & 0.02 & 0.42 \\
 & \textbf{\method} & \textbf{0.97} & \textbf{41.03} & \textbf{0.06} & \textbf{0.07} & \textbf{81.18} & \textbf{0.06} & \textbf{0.93} & \textbf{0.08} & 35.68 & \underline{0.47} & \underline{0.38} & \textbf{0.10} \\
\bottomrule
\end{tabular}
\caption{Comparison of protection methods on VGGFace2 and CelebA-HQ.
$\uparrow$/$\downarrow$ indicates higher/lower is better. Best in \textbf{bold}, second best \underline{underlined}.
P1: \textit{a photo of sks person}; P2: \textit{a dslr portrait of sks person}.}
\label{tab:main}
\end{table*}

\subsection{Setup}

\paragraph{Datasets.}
We evaluate \method on VGGFace2~\cite{cao2018vggface2} and CelebA-HQ~\cite{karras2018celeba}.
Following Anti-DreamBooth~\cite{le2023antidreambooth}, we select 50 identities from each dataset, using 4 images per identity as the target set to be protected.
Ablation studies and additional experiments use a 10-identity subset to reduce computational cost.
All images are resized to $512 \times 512$ for both perturbation generation and model fine-tuning.

\paragraph{Baselines.}
We compare against five representative protection methods:
Anti-DreamBooth~\cite{le2023antidreambooth} (Anti-DB), Mist~\cite{liang2023mist}, SDS(-)~\cite{xue2024sds}, MYOPIA~\cite{wu2025myopia}, and VCPro~\cite{mi2026vcpro}.
These baselines cover both established protection methods and recent quality-aware approaches.
For a fair comparison, we set $\epsilon=0.05$ for all methods except MYOPIA, whose original upper bound ($\epsilon=0.5$) is retained because its internal magnitude control prevents the realized perturbation from saturating this nominal budget.

\paragraph{Metrics.}
We evaluate along two dimensions: the image quality of protected images and the protection effectiveness against personalized generation.
For image quality, we measure the distortion between protected and clean images using
PSNR, SSIM~\cite{wang2004ssim}, LPIPS~\cite{zhang2018lpips}, and DISTS~\cite{ding2020dists}.
For protection effectiveness, we assess the identity fidelity and perceptual quality of DreamBooth-generated images using four metrics:
ISM is the ArcFace~\cite{deng2019arcface} identity similarity between generated and reference facial images, FDFR is the fraction of generated images for which no face is detected, SER-FIQ~\cite{terhorst2020serfiq} measures generated-face quality, and BRISQUE~\cite{mittal2012brisque} measures no-reference image quality.

\paragraph{Implementation details.}
We optimize the Laplace loss in Eq.~\eqref{eq:laplace} for $T{=}300$ iterations with sparsity ratio $r{=}0.05$.
The saliency map is obtained via BiSeNet~\cite{yu2018bisenet} face parsing.
We use Stable Diffusion v2.1~\cite{rombach2022ldm} as the target model and fine-tune with DreamBooth~\cite{ruiz2023dreambooth} for 1000 steps.
For each protected identity, we generate 32 images per evaluation prompt. 
The full prompt list is given in Appendix~\ref{app:setup}.
All experiments are conducted on NVIDIA A100 GPUs.
Full hyperparameters are provided in Appendix~\ref{app:implementation}.

\subsection{Main Results}

\paragraph{Quantitative results.}
Table~\ref{tab:main} compares all methods on VGGFace2 and CelebA-HQ under two inference prompts.
With only 5\% of pixels modified, \method achieves the best protected-image quality across all four metrics on both datasets.
Under P1, \method attains FDFR $=$ 0.93 and ISM $\leq$ 0.11, indicating a high face-detection failure rate and low target-identity similarity.
However, baselines face a pronounced trade-off: methods with dense perturbations introduce substantial artifacts, while those preserving image quality provide negligible identity disruption.
Under prompt P2, \method achieves the lowest ISM on both datasets ($\leq$ 0.15), indicating the lowest target-identity fidelity among the compared methods.
Extended results under other prompts are reported in Appendix~\ref{app:qualitative}.

\paragraph{Qualitative results.}
Figure~\ref{fig:qualitative} shows visual comparisons of protected images and their DreamBooth generations.
Our protected images are nearly indistinguishable from the originals: the sparse perturbation leaves 95\% of pixels untouched, and the saliency-guided dispersion penalty prevents visible clustering on facial regions.
Baselines, by contrast, introduce noticeable texture distortions or color shifts.
Although our protected images are barely altered, the faces in our generations are unrecognizable or cannot be detected, whereas baseline generations often still reveal the target identity. 
Additional qualitative comparisons are provided in Appendix~\ref{app:qualitative}.

\subsection{Uncontrolled Conditions}

In real-world scenarios, adversaries may use unknown models, obtain only a subset of protected images, or apply image pre-processing before fine-tuning.
We evaluate transferability and partial protection below; additional pre-processing results are provided in Appendix~\ref{app:robustness}.

\begin{figure}[!h]
\centering
\includegraphics[width=\columnwidth]{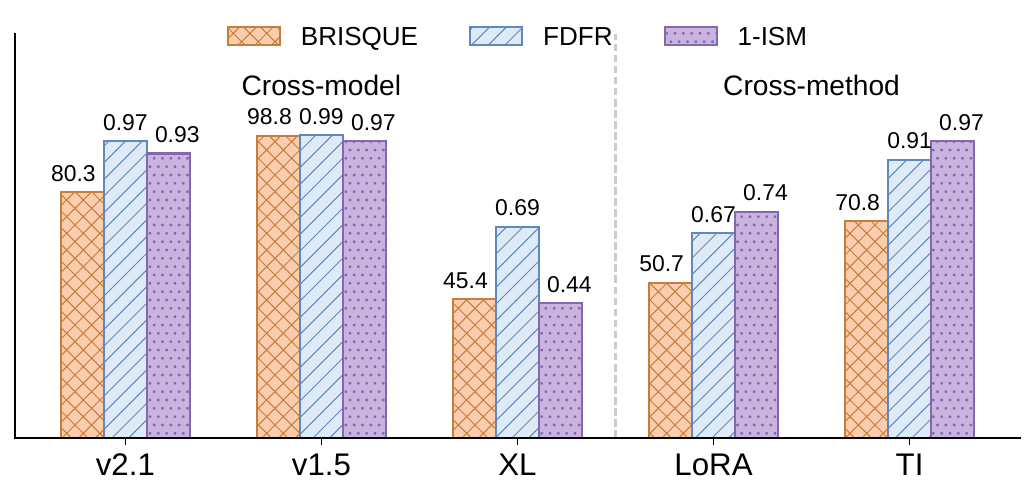}
\caption{Cross-model and cross-method transferability on CelebA-HQ. Bars are rescaled to a common range (taller $=$ better). Prompt: \textit{a photo of sks person}.}
\label{fig:transfer}
\end{figure}

\paragraph{Transferability.}
To assess cross-model transferability, we generate protected images using the VAE of SD~2.1 as the surrogate and test protection performance on different target models and personalization methods.
As shown in Figure~\ref{fig:transfer}, these results establish transferability across both target model and personalization changes.

\paragraph{Protected-image ratio.}
In practice, adversaries may collect images from diverse sources, mixing protected and unprotected ones.
We evaluate protection effectiveness when only a subset of the DreamBooth training images per identity are protected.
As shown in Table~\ref{tab:poisoning}, protection remains effective at 50\% (FDFR = 0.69).
Even at 25\%, our method still outperforms quality-oriented baselines under full protection (FDFR $\leq$ 0.02 in Table~\ref{tab:main}).

\begin{table}[!h]
\centering
\setlength{\tabcolsep}{3pt}
\begin{tabular}{c ccc}
\toprule
Ratio & BRISQUE $\uparrow$ & SER-FIQ $\downarrow$ & FDFR $\uparrow$ \\
\midrule
100\% & 67.64 & 0.07 & 0.93 \\
75\% & 59.78 & 0.12 & 0.88 \\
50\% & 48.71 & 0.31 & 0.69 \\
25\% & 30.51 & 0.59 & 0.33 \\
\bottomrule
\end{tabular}
\caption{Protection under different protected-image ratios on CelebA-HQ. Prompt: \textit{a photo of sks person}.}
\label{tab:poisoning}
\end{table}

\subsection{Efficiency Analysis}

Table~\ref{tab:efficiency} reports the wall-clock time and peak GPU memory required to protect one four-image identity set for \method and the baselines.
\method optimizes perturbations by backpropagating only through a frozen VAE encoder, yielding a lightweight protection procedure.
It requires approximately 70 seconds and 9 GB of GPU memory, corresponding to about one-sixth of the time and half the memory used by Anti-DB.
At a similar computational cost to SDS(-), \method achieves better protected-image quality and lower identity similarity on both datasets (Table~\ref{tab:main}).

\begin{table}[!h]
\centering
\begin{tabular}{l cc}
\toprule
Method & GPU Mem.\ (GB)  & Time (s) \\
\midrule
Anti-DB & ${\sim}$17 & ${\sim}$420 \\
Mist    & \underline{${\sim}$11} & ${\sim}$120 \\
SDS(-)  & \textbf{${\sim}$9}  & \textbf{${\sim}$70} \\
MYOPIA  & ${\sim}$18 & ${\sim}$590  \\
VCPro   & ${\sim}$20 & \underline{${\sim}$100}  \\
\method & \textbf{${\sim}$9}  & \textbf{${\sim}$70}  \\
\bottomrule
\end{tabular}
\caption{Efficiency comparison. Time and GPU memory are measured for protecting one four-image identity set at $512 \times 512$ resolution on a single NVIDIA A100.}
\label{tab:efficiency}
\end{table}

\subsection{Ablation Study}

Unless otherwise specified, all ablation experiments are conducted on CelebA-HQ with default hyperparameters.
Additional ablation studies on dispersion weight $\lambda$, density threshold $\tau$, and optimization steps $T$ are provided in Appendix~\ref{app:ablation}.

\begin{figure}[h]
\centering
\begin{subfigure}[b]{0.235\columnwidth}
  \centering
  \includegraphics[width=\linewidth]{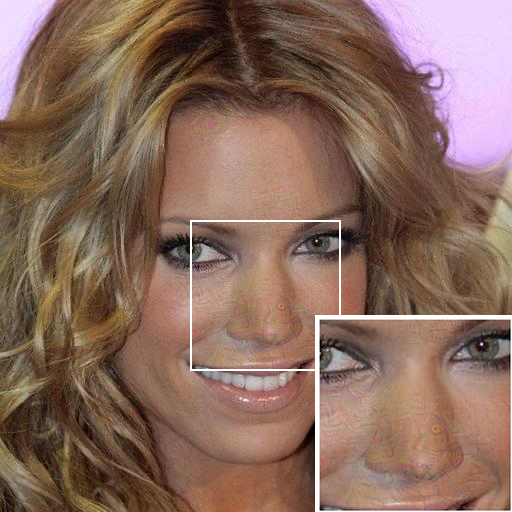}
  \caption{w/o Sal.}
\end{subfigure}
\hfill
\begin{subfigure}[b]{0.235\columnwidth}
  \centering
  \includegraphics[width=\linewidth]{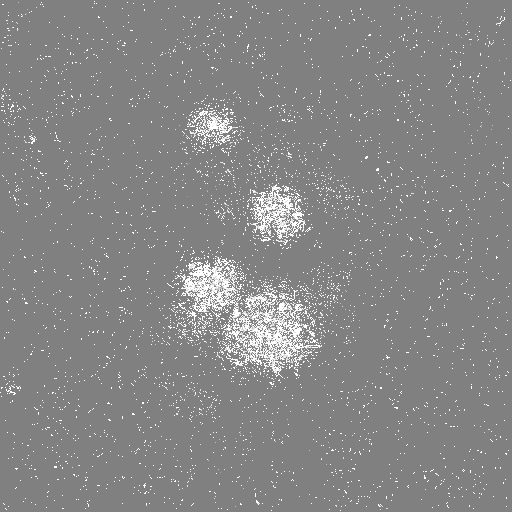}
  \caption{Mask}
\end{subfigure}
\hfill
\begin{subfigure}[b]{0.235\columnwidth}
  \centering
  \includegraphics[width=\linewidth]{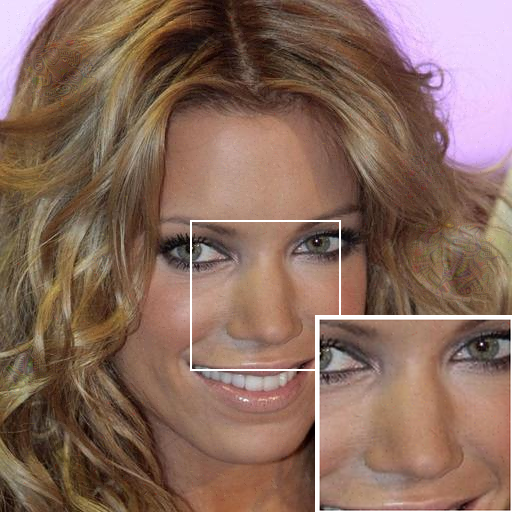}
  \caption{Full}
\end{subfigure}
\hfill
\begin{subfigure}[b]{0.235\columnwidth}
  \centering
  \includegraphics[width=\linewidth]{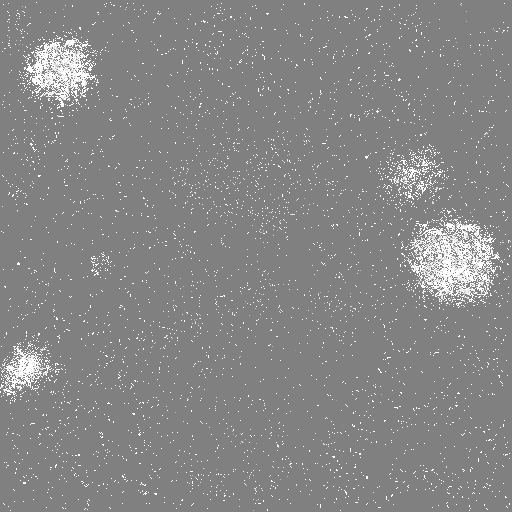}
  \caption{Mask}
\end{subfigure}
\caption{
Effect of the saliency-guided dispersion penalty.
(a-b) Without the penalty, selected locations concentrate around the eyes and nose.
(c-d) With the penalty, the selected locations are distributed more broadly across less salient regions.
}
\label{fig:saliency_effect}
\end{figure}

\paragraph{Component ablation.}
Table~\ref{tab:ablation_component} reports the results.
Replacing the learned top-$k$ mask with a random mask degrades ISM from 0.07 to 0.68, showing that learned mask placement is critical under a sparse budget.
The Laplace-based (LA) loss achieves ISM 0.07, outperforming the mean-squared-error (MSE) loss at 0.37 and the Lagrange Entropy (LE) loss~\cite{yang2025variance} at 0.25; both alternative objectives are defined in Appendix~\ref{app:loss_variants}.
Removing the saliency penalty improves protection, reducing ISM to 0.02, while leaving global LPIPS nearly unchanged.
Since global LPIPS can obscure local distortion, we additionally report F-LPIPS: a facial mask is obtained by applying SAM~\cite{kirillov2023sam} to the clean image and then applied to the protected image to compute mean LPIPS within the facial region.
The lower F-LPIPS of the full model shows that the saliency term reduces facial distortion by dispersing perturbations away from salient regions, at a modest cost in protection strength.
As Figure~\ref{fig:saliency_effect} shows, the mask without the dispersion penalty concentrates around the eyes and nose, while the full model distributes the selected locations more broadly across less salient regions.

\begin{table}[!h]
\centering
\small
\setlength{\tabcolsep}{2pt}
\begin{tabular}{l ccc | ccc}
\toprule
 & Top-$k$ & Saliency &  LA & F-LPIPS$\downarrow$ & LPIPS $\downarrow$ & ISM $\downarrow$ \\
\midrule
w/o Saliency    & $\checkmark$ &              & $\checkmark$ & 0.05 & 0.06 & 0.02 \\
Random Mask     &              & $\checkmark$ & $\checkmark$ & 0.03 & 0.05 & 0.68 \\
MSE Loss        & $\checkmark$ & $\checkmark$ &              & 0.04 & 0.07 & 0.37 \\
LE Loss         & $\checkmark$ & $\checkmark$ &              & 0.03 & 0.06 & 0.25 \\
Full model      & $\checkmark$ & $\checkmark$ & $\checkmark$ & 0.03 & 0.07 & 0.07 \\
\bottomrule
\end{tabular}
\caption{Component ablation on CelebA-HQ. Prompt: \textit{a photo of sks person}.}
\label{tab:ablation_component}
\end{table}

\begin{figure}[!h]
\centering
\includegraphics[width=\columnwidth]{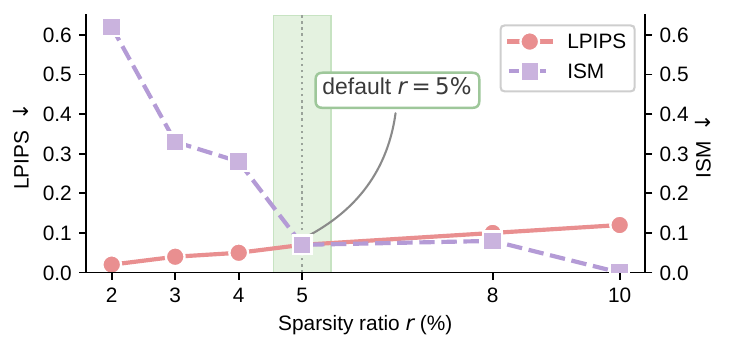}
\caption{Trade-off between protected-image distortion (LPIPS) and protection effectiveness (ISM) across sparsity ratios. The default setting $r{=}5\%$ provides a balanced operating point. Prompt: \textit{a photo of sks person}.}
\label{fig:ratio_tradeoff}
\end{figure}

\paragraph{Sparsity ratio.}
We vary the sparsity ratio $r$ from 2\% to 10\% to investigate the trade-off between image quality and protection strength.
Figure~\ref{fig:ratio_tradeoff} visualizes the trade-off between visibility cost (LPIPS) and protection effectiveness (ISM) and full metrics are provided in Table~\ref{tab:ablation_sparsity} in the appendix.
At $r{=}5\%$, ISM drops to 0.07 while LPIPS remains at only 0.07, indicating an effective operating point that balances both protection effectiveness and protected-image quality.

%% file: sections/06_Conclusion.tex
\section{Conclusion}

In this paper, we propose \method, an $L_0$-constrained sparse perturbation framework for face privacy protection.
\method modifies only a small fraction of pixels and targets the VAE encoder's latent variance to achieve strong identity disruption under a tight perturbation budget.
A saliency-guided dispersion penalty further suppresses perceptible concentration in perceptually salient facial regions.
Comprehensive experiments on CelebA-HQ and VGGFace2 demonstrate that, with only 5\% of pixels perturbed, \method attains the best protected-image quality and the lowest identity similarity among the compared baselines.
By optimizing where the sparse perturbation lands, \method achieves strong protection and high image quality together.

\paragraph{Limitations.}
\method targets fine-tuning-based personalization, in which the protected images pass through the VAE encoder during training.
Like other protections that target the VAE encoder, it does not affect encoder-based personalization methods such as IP-Adapter FaceID~\cite{ye2023ipadapter}.
Smoothing pre-processing, such as JPEG compression, Gaussian blur, and resizing, removes most of the sparse perturbation and largely restores identity similarity (Table~\ref{tab:robustness} in the appendix), whereas mild additive noise is tolerated.
This vulnerability is shared by pixel-level protective perturbations in general~\cite{sun2026protect}, and improving robustness to such transformations is left for future work.

%% file: sections/08_Appendix.tex
\section*{Appendix Overview}
This appendix provides the algorithmic, experimental, empirical details and reports additional results that further support our findings. 
It is organized as follows:

\begin{itemize}
    \item Appendix~\ref{app:algorithm}: the complete optimization algorithm.
    \item Appendix~\ref{app:setup}: the experimental setup, including datasets, baselines, implementation details, the generation protocol and evaluation metrics.
    \item Appendix~\ref{app:experiments}: additional experiments, including ablation studies, robustness to image pre-processing, cross-model and cross-method transferability, and qualitative results.
\end{itemize}

\section{Algorithm}
\label{app:algorithm}
Algorithm~\ref{alg:ours} details the complete optimization procedure of \method.
\begin{algorithm}[h]
\caption{\method}
\label{alg:ours}
\begin{algorithmic}[1]
\REQUIRE Image $x$, VAE encoder $\mathcal{E}_\theta$, perturbation budget $\epsilon$, sparsity ratio $r$, step sizes $\alpha, \beta$, dispersion weight $\lambda$, density threshold $\tau$, kernel size $K$, max steps $T$, patience $P$
\STATE $k \leftarrow \lfloor r \cdot H \cdot W \rfloor$
\STATE Initialize $\delta \sim \mathrm{Uniform}(-\epsilon, \epsilon)$, $\ell \sim \mathcal{N}(0, 1)$
\STATE Compute saliency map $S \leftarrow \mathrm{BiSeNet}(x)$
\STATE Build kernel $\mathbf{W} \leftarrow \frac{1}{K^2}\mathbf{1}_{K \times K}$
\STATE $p \leftarrow 0$; $m_\mathrm{prev} \leftarrow \mathbf{0}$; $x^\star \leftarrow x$; $J^\star \leftarrow -\infty$
\FOR{$t = 1$ to $T$}
  \STATE $\hat{m} \leftarrow \mathrm{sig}(\ell)$; $m \leftarrow \mathrm{TopK}(\hat{m}, k)$
  \STATE $x' \leftarrow \mathrm{clip}(x + \delta \odot m,\, 0,\, 1)$
  \STATE $(\mu', {\sigma'}^2) \leftarrow \mathcal{E}_\theta(x')$
  \STATE Compute $\mathcal{L}_\mathrm{var}$ via Eq.~\eqref{eq:laplace}
  \STATE Compute $\Omega$ via Eq.~\eqref{eq:dispersion_penalty} using STE on $m$
  \STATE $J \leftarrow \mathcal{L}_\mathrm{var} - \lambda\,\Omega$
  \IF{$J > J^\star$}
    \STATE $J^\star \leftarrow J$;\ \ $x^\star \leftarrow x'$
  \ENDIF
  \STATE $\delta \leftarrow \Pi_{[-\epsilon,\epsilon]}(\delta + \alpha \cdot \mathrm{sign}(\nabla_\delta \mathcal{L}_\mathrm{var}))$
  \STATE $g_\ell \leftarrow$ balanced direction via Eq.~\eqref{eq:grad_balance}
  \STATE $\ell \leftarrow \ell + \sqrt{HW}\,\beta \cdot g_\ell$ 
  \IF{$m = m_\mathrm{prev}$}
    \STATE $p \leftarrow p + 1$
  \ELSE
    \STATE $p \leftarrow 0$
  \ENDIF
  \IF{$p \geq P$}
    \STATE Reinitialize $\delta$, $\ell$; $p \leftarrow 0$
  \ENDIF
  \STATE $m_\mathrm{prev} \leftarrow m$
\ENDFOR
\RETURN $x^\star$, the iterate with the highest $\mathcal{L}_\mathrm{var}-\lambda\Omega$
\end{algorithmic}
\end{algorithm}

\section{Experimental Setup}
\label{app:setup}

\subsection{Datasets}
We conduct experiments on two widely used face datasets, CelebA-HQ~\cite{karras2018celeba} and VGGFace2~\cite{cao2018vggface2}, following prior privacy-protection work~\cite{le2023antidreambooth,mi2026vcpro}.
CelebA-HQ is a high-quality version of CelebA consisting of 30{,}000 celebrity face images, while VGGFace2 is a large-scale dataset with over 3.3 million images from 9{,}131 identities.
From each dataset we select 50 identities for the main comparison and 10 identities for additional experiments; for each identity we use 4 images as the target set to be protected and reserve an additional set of clean images as identity references for evaluation.
\subsection{Baselines}
We compare against five representative protection methods that span adversarial optimization against DreamBooth and quality-aware protection:
Anti-DreamBooth~\cite{le2023antidreambooth}, Mist~\cite{liang2023mist}, SDS(-)~\cite{xue2024sds}, MYOPIA~\cite{wu2025myopia}, and VCPro~\cite{mi2026vcpro}.
For a fair comparison, we set the $L_\infty$ perturbation budget to $\epsilon =0.05$ for all methods except MYOPIA, for which we retain its original nominal bound ($\epsilon = 0.5$) together with its JND-based magnitude control.
The remaining hyperparameters follow each method's official implementation:
\begin{itemize}\setlength\itemsep{1pt}
  \item \textbf{Mist} and \textbf{SDS(-)}: 100 optimization steps with PGD step size $\alpha = 1$.
  \item \textbf{VCPro}: 1000 adversarial optimization steps with PGD step size $\alpha = 1$.
  \item \textbf{Anti-DreamBooth}: 50 training steps, with 3 model fine-tuning steps and 6 adversarial PGD steps, a checkpointing interval of 10, a surrogate learning rate of $5 \times 10^{-7}$, and PGD step size $5 \times 10^{-3}$.
  \item \textbf{MYOPIA}: 10 training steps, with 3 model fine-tuning steps and 100 adversarial PGD steps, a checkpointing interval of 10, a surrogate learning rate of $5 \times 10^{-7}$, and PGD step size $5 \times 10^{-3}$.
\end{itemize}

\subsection{Implementation Details}
\label{app:implementation}

\paragraph{Perturbation optimization.}
The $L_\infty$ bound on individual perturbed pixels is $\epsilon = 0.05$, with a PGD step size of $\alpha = 0.25$.
The default sparsity ratio $r = 0.05$ yields an $L_0$ budget of $k = \lfloor 0.05 \times 512 \times 512 \rfloor = 13{,}107$ pixels.
The mask logits are updated with step size $\beta = 0.25$, and the patience-based reinitialization threshold is $P = 3$.
For the saliency-guided dispersion penalty, the local density is estimated with a $7 \times 7$ mean-pooling kernel, the density threshold is $\tau = 0.15$, and the dispersion weight is $\lambda = 0.5$.

\paragraph{Saliency class weights.}
The saliency map $S(x)$ is obtained from BiSeNet face parsing~\cite{yu2018bisenet}, which labels each pixel with one of 19 semantic classes.
Each class is assigned a fixed perceptual weight $w \in [0, 1]$, with the complete class-to-weight mapping listed in Table~\ref{tab:saliency_weights}.
\begin{table}[h]
\centering
\setlength{\tabcolsep}{4pt}
\begin{tabular}{clc@{\hskip 12pt}clc}
\toprule
ID & Class & $w$ & ID & Class & $w$ \\
\midrule
0 & Background  & 0.00 & 10 & Nose      & 0.90 \\
1 & Skin        & 0.80 & 11 & Mouth     & 0.90 \\
2 & L.\ eyebrow & 0.70 & 12 & Upper lip & 0.90 \\
3 & R.\ eyebrow & 0.70 & 13 & Lower lip & 0.90 \\
4 & L.\ eye     & 1.00 & 14 & Neck      & 0.50 \\
5 & R.\ eye     & 1.00 & 15 & Necklace  & 0.10 \\
6 & Eyeglasses  & 0.60 & 16 & Cloth     & 0.05 \\
7 & L.\ ear     & 0.10 & 17 & Hair      & 0.02 \\
8 & R.\ ear     & 0.10 & 18 & Hat       & 0.10 \\
9 & Earring     & 0.10 &    &           &      \\
\bottomrule
\end{tabular}
\caption{Complete BiSeNet class-to-weight mapping used by the saliency-guided dispersion penalty. Higher weights $w$ more strongly penalize perturbation concentration in that region.}
\label{tab:saliency_weights}
\end{table}

\paragraph{DreamBooth fine-tuning.}
We fine-tune with DreamBooth using prior preservation: the instance prompt is ``a photo of \textit{sks} person'' and the class (prior) prompt is ``a photo of person'', with 200 class images and a prior-loss weight of $1.0$.
Training uses a learning rate of $5 \times 10^{-7}$, a batch size of 2, 1000 steps, text-encoder training enabled, and bf16 precision.
All images are resized to $512 \times 512$.

\subsection{Generation Protocol}
For every method, subject, and prompt, personalized images are generated using the \texttt{DDPMScheduler} shipped with the SD~2.1-base checkpoint, with 100 inference steps, a guidance scale of $7.5$, a resolution of $512 \times 512$, and shared random seeds across methods.
Evaluation uses the following six prompts, referenced as P1 to P6:
\begin{itemize}\setlength\itemsep{1pt}
  \item \textbf{P1}: ``a photo of \textit{sks} person''
  \item \textbf{P2}: ``a dslr portrait of \textit{sks} person''
  \item \textbf{P3}: ``a photo of \textit{sks} person looking at the mirror''
  \item \textbf{P4}: ``a photo of \textit{sks} person in front of eiffel tower''
  \item \textbf{P5}: ``a close-up photo of \textit{sks} person, high details''
  \item \textbf{P6}: ``a photo of \textit{sks} person stand in front of the table holding a knife''
\end{itemize}

\subsection{Evaluation Metrics}
We report metrics along two dimensions: the image quality of the protected image and the protection effectiveness against personalized generation.

\emph{Image quality} measures the distortion between a protected image and its clean counterpart.
PSNR and SSIM~\cite{wang2004ssim} quantify pixel- and structure-level fidelity (higher is better), whereas LPIPS~\cite{zhang2018lpips} and DISTS~\cite{ding2020dists} measure deep perceptual and structure--texture distance (lower is better).

\emph{Protection effectiveness} evaluates the DreamBooth-generated images in terms of identity and perceptual quality.
ISM (Identity Score Metric) is the ArcFace~\cite{deng2019arcface} cosine similarity between generated and reference faces; a lower value means less of the target identity is reproduced.
FDFR (Face Detection Failure Rate) is the fraction of generated images in which a face detector fails to locate any face; a higher value indicates stronger protection.
SER-FIQ~\cite{terhorst2020serfiq} is a no-reference face image quality score, for which a lower value reflects less recognizable generated faces.
BRISQUE~\cite{mittal2012brisque} is a no-reference image quality metric, for which higher values indicate more artifacts in the generated images.

\section{Additional Experiments and Results}
\label{app:experiments}

\subsection{Loss Function}
\label{app:loss_variants}

The component ablation compares our default optimization objective in Eq.~\eqref{eq:laplace} against two alternative variance-disruption losses, all evaluated within the same sparse framework (top-$k$ mask, saliency penalty, $r = 0.05$).

\paragraph{MSE loss.}
This variant maximizes the squared deviation between the latent variance of the clean and protected images:
\begin{equation}
\mathcal{L}_\mathrm{MSE} = \sum_{j} \left( \sigma_{\mathrm{clean},j}^2 - \sigma_j^2 \right)^2,
\label{eq:mse_loss}
\end{equation}
where $\sigma_{\mathrm{clean},j}^2$ and $\sigma_j^2$ are the $j$-th latent variance components of the clean image $x$ and the protected image $x'$, respectively.

\paragraph{Lagrange Entropy (LE) loss.}
Following~\cite{yang2025variance}, the LE loss combines an entropy term on the latent variance with a soft normalization constraint:
\begin{equation}
\mathcal{L}_\mathrm{LE} = -\sum_{j} \sigma_j^2 \log\!\left( \sigma_j^2 + \varepsilon \right) + \lambda_c \left( \sum_{j} \sigma_j^2 - 1 \right)^{2},
\label{eq:le_loss}
\end{equation}
where $\varepsilon = 10^{-8}$ ensures numerical stability and $\lambda_c = 0.1$ weights the constraint that softly normalizes the total latent variance.

\subsection{Additional Ablation Studies}
\label{app:ablation}

\paragraph{Sparsity ratio.}
Table~\ref{tab:ablation_sparsity} reports the full metrics across sparsity ratios.
Image quality degrades approximately linearly with $r$: PSNR decreases by roughly 2\,dB for every 2\% increase in the ratio, and SSIM drops from 0.99 at $r{=}2\%$ to 0.93 at $r{=}10\%$.
Protection, by contrast, exhibits a pronounced phase transition between $r{=}4\%$ and $r{=}5\%$: ISM drops from 0.28 to 0.07, FDFR jumps from 0.73 to 0.97, and SER-FIQ falls from 0.26 to 0.06, indicating that 5\% of spatial locations is the critical budget at which the sparse perturbation becomes sufficient to disrupt identity acquisition.

\begin{table}[h]
\centering
\small
\setlength{\tabcolsep}{2pt}
\begin{tabular}{c cccc cccc}
\toprule
 & \multicolumn{2}{c}{Image Quality} & \multicolumn{4}{c}{Protection} \\
\cmidrule(lr){2-3} \cmidrule(lr){4-7}
$r$ (\%) & SSIM$\uparrow$ & LPIPS$\downarrow$ & BRISQUE$\uparrow$ & SER-FIQ$\downarrow$ & FDFR$\uparrow$ & ISM$\downarrow$ \\
\midrule
2   & 0.99  & 0.02 &  26.68 & 0.67 & 0.16 & 0.62 \\
3   & 0.98  & 0.04 &  43.62 & 0.33 & 0.61 & 0.33 \\
4   & 0.98  & 0.05 &  53.26 & 0.26 & 0.73 & 0.28 \\
5   & 0.97  & 0.07 &  80.27 & 0.06 & 0.97 & 0.07 \\
8   & 0.95  & 0.10 &  90.93 & 0.12 & 0.85 & 0.08 \\
10  & 0.93  & 0.12 &  121.43 & 0.01 & 1.00 & 0.00 \\
\bottomrule
\end{tabular}
\caption{Effect of sparsity ratio $r$ on CelebA-HQ. Prompt: \textit{a photo of sks person}.}
\label{tab:ablation_sparsity}
\end{table}

\begin{figure}[h]
\centering
\begin{subfigure}[b]{\columnwidth}
  \centering
  \includegraphics[width=\linewidth]{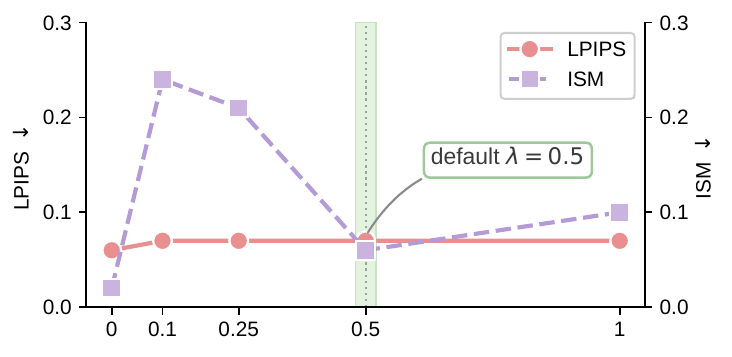}
  \caption{Dispersion weight $\lambda$}
\end{subfigure}

\begin{subfigure}[b]{\columnwidth}
  \centering
  \includegraphics[width=\linewidth]{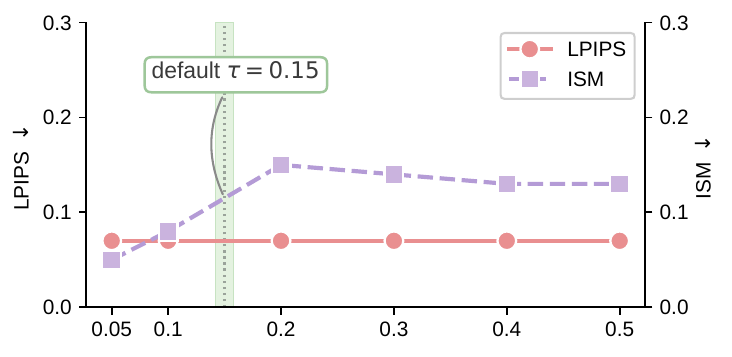}
  \caption{Density threshold $\tau$}
\end{subfigure}

\begin{subfigure}[b]{\columnwidth}
  \centering
  \includegraphics[width=\linewidth]{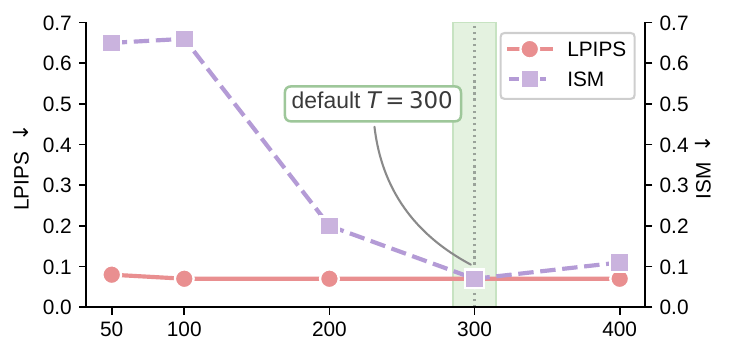}
  \caption{Optimization steps $T$}
\end{subfigure}
\caption{%
Hyperparameter sensitivity on CelebA-HQ (prompt: \textit{a photo of sks person}).
LPIPS ($\downarrow$, perceptual cost) and ISM ($\downarrow$, protection) versus (a) the dispersion weight $\lambda$, (b) the density threshold $\tau$, and (c) the number of optimization steps $T$; the default setting is highlighted in each panel.}
\label{fig:sensitivity}
\end{figure}

\paragraph{Hyperparameter sensitivity.}
Figure~\ref{fig:sensitivity} shows the LPIPS--ISM trade-off as we vary the dispersion weight $\lambda$, the density threshold $\tau$, and the number of optimization steps $T$.
LPIPS (perceptual cost) stays nearly flat throughout, so tuning these hyperparameters affects protection (ISM) at almost no perceptual cost.
The dispersion weight $\lambda$ shows a non-monotonic trend: $\lambda = 0$ (no penalty) yields the strongest raw protection because the mask freely clusters on the most effective locations, but a weak penalty ($\lambda = 0.1$) disturbs mask optimization without a clear dispersion signal and weakens protection, which recovers for $\lambda \geq 0.5$.
We adopt $\lambda = 0.5$, which restores near-maximal protection while dispersing perturbations away from salient regions; the spatial effect is visualized in Fig.~\ref{fig:saliency_effect} of the main text.
Protection is largely insensitive to $\tau$ over $[0.05, 0.5]$, while ISM falls sharply between $T = 100$ and $T = 300$ and changes little beyond it, so we set $T = 300$ as the default.

\subsection{Uncontrolled Conditions}
\label{app:robustness}

\begin{figure*}[t]
\centering
\includegraphics[width=\textwidth]{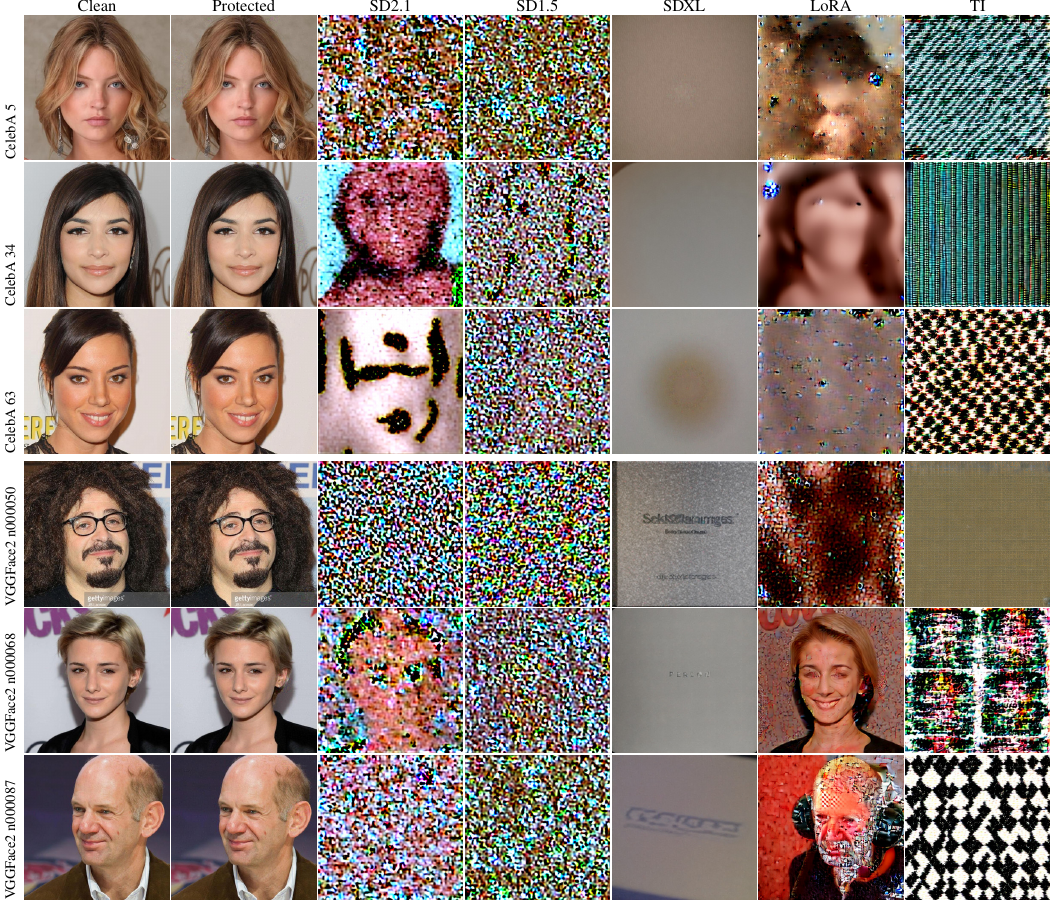}
\caption{%
Cross-model and cross-method transferability on CelebA-HQ and VGGFace2.
Columns: clean image, our protected image, and personalized generations under the surrogate (SD~2.1), two unseen backbones (SD~1.5, SDXL), and two unseen personalization methods (LoRA, Textual Inversion).
All generations use the prompt \textit{``a photo of sks person''}.
The protected identity fails to be reproduced across all settings.}
\label{fig:transfer_qual}
\end{figure*}

\paragraph{Transferability.}
\label{app:transfer_qual}
Complementing the quantitative transfer results in the main text (Fig.~\ref{fig:transfer}), Fig.~\ref{fig:transfer_qual} shows qualitative generations when the protected images, optimized on the SD~2.1 surrogate, are used to personalize \emph{unseen} backbones and methods.
Across cross-backbone (SD~1.5, SDXL) and cross-method (LoRA, Textual Inversion) settings, the target identity is never faithfully reconstructed, confirming that our protection does not overfit to the surrogate model.
Table~\ref{tab:transfer} quantifies this: protection transfers strongly to the SD~1.5 backbone and to the LoRA and Textual Inversion personalizers (ISM $\leq 0.26$).
SDXL is the strictest of these settings, as it replaces the SD~2.1 encoder with a retrained VAE operating at a different latent scale; even there, $69\%$ of the generations contain no detectable face.

\begin{table}[t]
\centering
\setlength{\tabcolsep}{5pt}
\begin{tabular}{ll cc}
\toprule
Setting & Target & FDFR$\uparrow$ & ISM$\downarrow$ \\
\midrule
Surrogate & SD~2.1 & 0.97 & 0.07 \\
\midrule
\multirow{2}{*}{Cross-backbone} & SD~1.5 & 0.99 & 0.03 \\
 & SDXL & 0.69 & 0.56 \\
\midrule
\multirow{2}{*}{Cross-method} & LoRA & 0.67 & 0.26 \\
 & Textual Inversion & 0.91 & 0.03 \\
\bottomrule
\end{tabular}
\caption{Quantitative cross-model and cross-method transferability on CelebA-HQ (prompt: \textit{a photo of sks person}). Protection is optimized on the SD~2.1 surrogate and evaluated on unseen backbones and personalization methods.}
\label{tab:transfer}
\end{table}

\paragraph{Image Pre-processing.}
We evaluate common image pre-processing that an adversary or platform might apply before fine-tuning.
Table~\ref{tab:robustness} reports results on VGGFace2 under JPEG compression, Gaussian blur, additive Gaussian noise, downsampling round-trip, and center crop-resize, for both our default setting and the denser $\lambda{=}0$ variant.
Additive noise, which does not selectively erase the sparse perturbation, is tolerated well (default ISM $=0.13$ at $\sigma{=}2$), whereas smoothing transforms (JPEG, blur, resize) remove part of the localized signal.
The $\lambda{=}0$ variant attains stronger raw protection (no transform: ISM $0.00$ vs.\ $0.09$) and slightly better noise tolerance, but under smoothing transforms both settings converge, reflecting a general trade-off between imperceptibility and robustness rather than an effect specific to $\lambda$.
This degradation under smoothing is consistent with prior findings that pixel-level protective perturbations are broadly vulnerable to real-world image transformations~\cite{sun2026protect}, and improving such robustness remains an open direction.

\begin{table*}[h]
\centering
\small
\setlength{\tabcolsep}{4pt}
\begin{tabular}{ll ccc ccc}
\toprule
 & & \multicolumn{3}{c}{Default} & \multicolumn{3}{c}{$\lambda=0$} \\
\cmidrule(lr){3-5} \cmidrule(lr){6-8}
Transform & Param & BRISQUE$\uparrow$ & FDFR$\uparrow$ & ISM$\downarrow$ & BRISQUE$\uparrow$ & FDFR$\uparrow$ & ISM$\downarrow$ \\
\midrule
None & -- & 82.34 & 0.89 & 0.09 & 102.16 & 1.00 & 0.00 \\
\midrule
\multirow{3}{*}{JPEG}
 & $q=85$ & 15.15 & 0.03 & 0.66 & 14.33 & 0.02 & 0.63 \\
 & $q=75$ & 17.39 & 0.03 & 0.66 & 17.21 & 0.03 & 0.62 \\
 & $q=55$ & 22.39 & 0.03 & 0.66 & 22.41 & 0.03 & 0.65 \\
\midrule
\multirow{3}{*}{Blur}
 & $r=1$ & 37.55 & 0.08 & 0.67 & 37.03 & 0.10 & 0.65 \\
 & $r=2$ & 57.69 & 0.05 & 0.66 & 57.92 & 0.06 & 0.66 \\
 & $r=3$ & 69.22 & 0.05 & 0.64 & 68.32 & 0.05 & 0.65 \\
\midrule
\multirow{3}{*}{Gauss.\ noise}
 & $\sigma=2$ & 62.92 & 0.92 & 0.13 & 74.83 & 1.00 & 0.00 \\
 & $\sigma=4$ & 42.47 & 0.51 & 0.35 & 52.15 & 0.77 & 0.24 \\
 & $\sigma=8$ & 6.05 & 0.03 & 0.66 & 5.44 & 0.02 & 0.64 \\
\midrule
Resize & $s=0.5$ & 44.57 & 0.08 & 0.67 & 43.78 & 0.08 & 0.66 \\
Crop   & $\rho=0.9$ & 23.94 & 0.07 & 0.64 & 22.62 & 0.08 & 0.64 \\
\bottomrule
\end{tabular}
\caption{Image pre-processing on VGGFace2 for the default and $\lambda=0$ settings. Prompt: \textit{a photo of sks person}.}
\label{tab:robustness}
\end{table*}

\subsection{Additional Quantitative and Qualitative Results}
\label{app:qualitative}

\begin{figure*}[p]
\centering
\includegraphics[width=\textwidth,height=0.92\textheight,keepaspectratio]{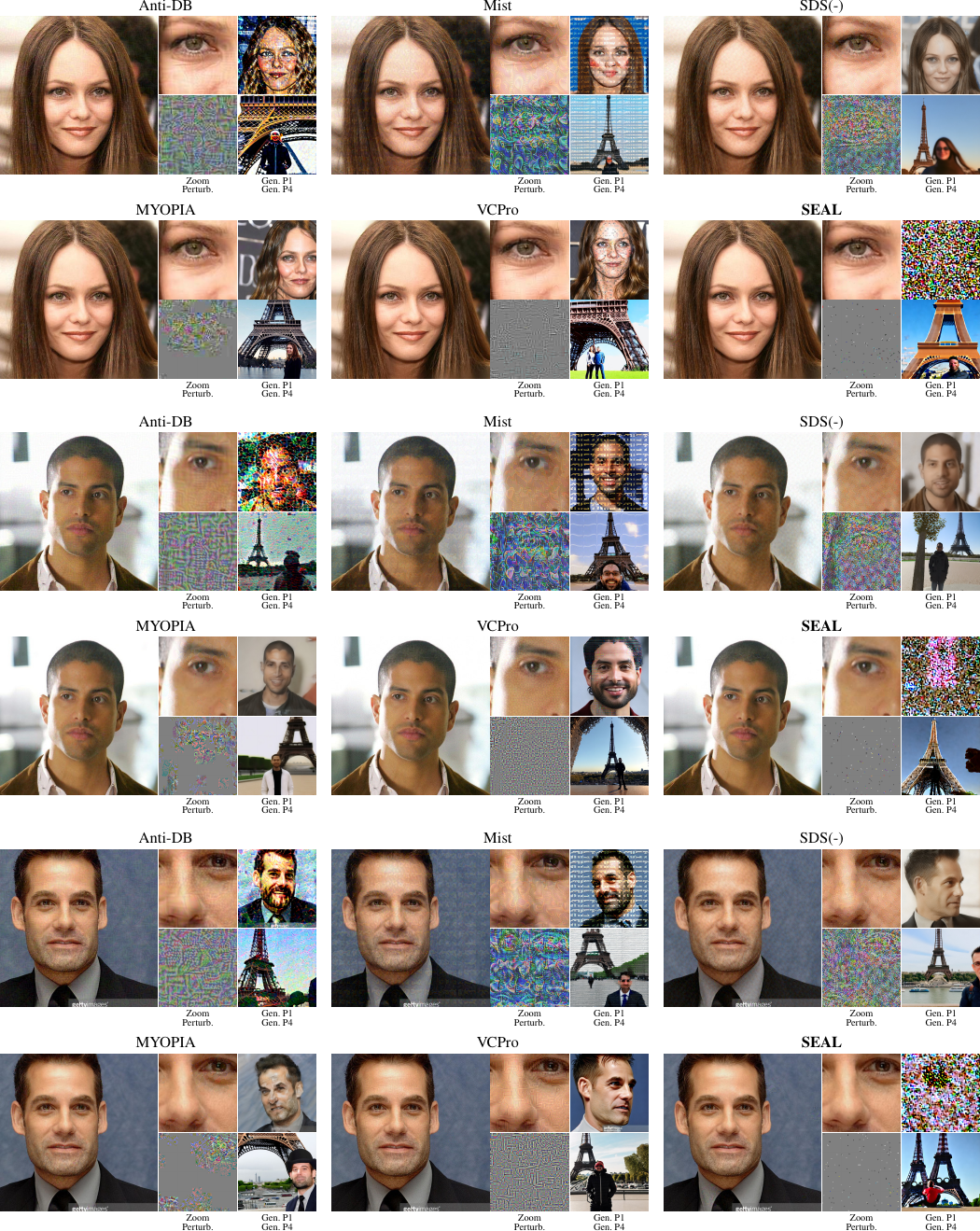}
\caption{%
Per-method qualitative comparison on three identities.
For each method we show the protected image, a zoomed crop, the perturbation residual, and two DreamBooth generations: Gen~P1 (\textit{``a photo of sks person''}) and Gen~P4 (\textit{``a photo of sks person in front of eiffel tower''}).
Compared with baselines, our perturbation is the least visible in the zoom and residual while still preventing identity reproduction.}
\label{fig:qual_percmethod}
\end{figure*}

\begin{figure*}[p]
\centering
\includegraphics[width=\textwidth,height=0.92\textheight,keepaspectratio]{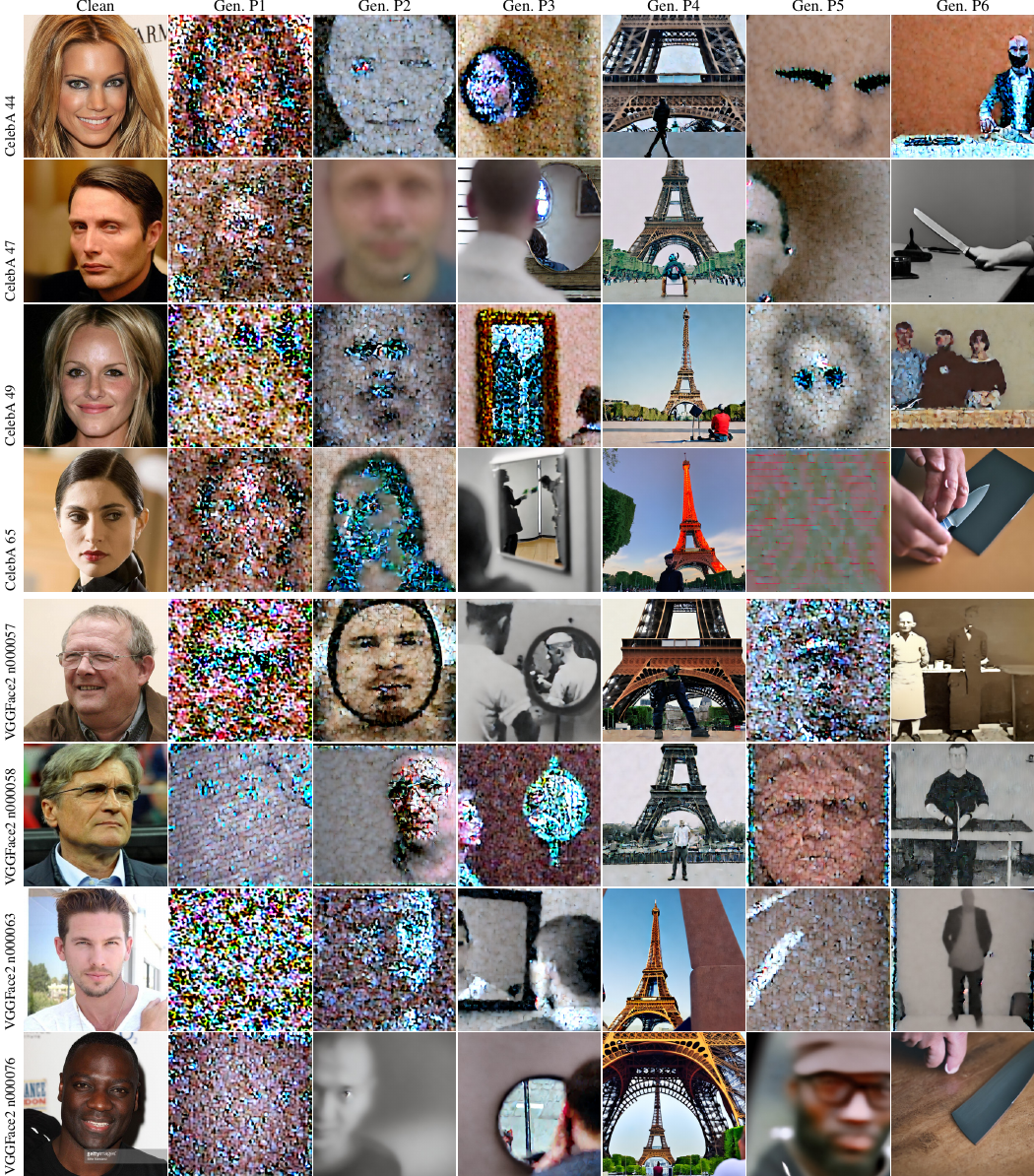}
\caption{%
Additional qualitative results of our method across identities from CelebA-HQ and VGGFace2.
Columns: clean image, and DreamBooth generations under six prompts. P1: \textit{``a photo of sks person''}; P2: \textit{``a dslr portrait of sks person''}; P3: \textit{``a photo of sks person looking at the mirror''}; P4: \textit{``a photo of sks person in front of eiffel tower''}; P5: \textit{``a close-up photo of sks person, high details''}; P6: \textit{``a photo of sks person stand in front of the table holding a knife''}. Protection holds consistently across identities and prompts.}
\label{fig:qual_grid}
\end{figure*}

Table~\ref{tab:prompt3456} extend the main comparison (Table~\ref{tab:main}) to prompts P3--P6, reporting protection effectiveness on both datasets.
Fig.~\ref{fig:qual_percmethod} compares our method against baselines on three identities: for each method we show the protected image, a zoomed crop, the perturbation residual, and two DreamBooth generations.
Our perturbation is the least conspicuous in the zoom and residual, yet the generated identity is still fully disrupted.
Fig.~\ref{fig:qual_grid} further illustrates the breadth of our protection across eight identities from CelebA-HQ and VGGFace2 under six evaluation prompts.

\begin{table*}[t]
\centering
\small
\setlength{\tabcolsep}{2pt}
\begin{tabular}{ll cccc cccc}
\toprule
 & & \multicolumn{4}{c}{Protection (P3)}
 & \multicolumn{4}{c}{Protection (P4)} \\
\cmidrule(lr){3-6} \cmidrule(lr){7-10}
Dataset & Method
 & BRISQUE$\uparrow$ & SER-FIQ$\downarrow$ & FDFR$\uparrow$ & ISM$\downarrow$
 & BRISQUE$\uparrow$ & SER-FIQ$\downarrow$ & FDFR$\uparrow$ & ISM$\downarrow$ \\
\midrule
\multirow{7}{*}{VGGFace2}
 & No Def.    & 20.26 & 0.58 & 0.02 & 0.42 & 22.14 & 0.53 & 0.26 & 0.22 \\
 & Anti-DB       & 35.44 & 0.27 & 0.36 & 0.24 & \textbf{39.26} & \underline{0.28} & 0.43 & \underline{0.11} \\
 & Mist          & 27.19 & \underline{0.26} & \underline{0.48} & \underline{0.12} & 28.04 & 0.36 & \textbf{0.66} & \textbf{0.09} \\
 & SDS(-)        & \underline{46.72} & 0.58 & 0.07 & 0.40 & 23.40 & 0.51 & 0.23 & 0.19 \\
 & MYOPIA        & \textbf{50.93} & 0.59 & 0.04 & 0.40 & \underline{36.44} & 0.45 & 0.38 & 0.19 \\
 & VCPro         & 17.10 & 0.58 & 0.03 & 0.42 & 22.43 & 0.55 & 0.18 & 0.21 \\
 & \textbf{\method} & 39.86 & \textbf{0.23} & \textbf{0.50} & \textbf{0.10} & 21.51 & \textbf{0.24} & \underline{0.54} & \textbf{0.09} \\
\midrule
\multirow{7}{*}{CelebA-HQ}
 & No Def.    & 19.27 & 0.58 & 0.03 & 0.44 & 23.46 & 0.54 & 0.22 & 0.16 \\
 & Anti-DB       & 37.49 & \underline{0.27} & \underline{0.46} & 0.21 & \textbf{39.75} & \underline{0.28} & 0.45 & \underline{0.07} \\
 & Mist          & 27.75 & 0.33 & 0.24 & \underline{0.13} & 30.85 & 0.33 & \textbf{0.68} & \underline{0.07} \\
 & SDS(-)        & \underline{41.62} & 0.58 & 0.08 & 0.36 & 21.04 & 0.47 & 0.38 & 0.09 \\
 & MYOPIA        & \textbf{53.15} & 0.62 & 0.06 & 0.40 & \underline{37.68} & 0.41 & 0.52 & 0.11 \\
 & VCPro         & 17.18 & 0.57 & 0.05 & 0.41 & 23.62 & 0.52 & 0.24 & 0.13 \\
 & \textbf{\method} & 37.76 & \textbf{0.21} & \textbf{0.56} & \textbf{0.08} & 20.75 & \textbf{0.24} & \underline{0.54} & \textbf{0.05} \\
\midrule
 & & \multicolumn{4}{c}{Protection (P5)}
 & \multicolumn{4}{c}{Protection (P6)} \\
\cmidrule(lr){3-6} \cmidrule(lr){7-10}
 &
 & BRISQUE$\uparrow$ & SER-FIQ$\downarrow$ & FDFR$\uparrow$ & ISM$\downarrow$
 & BRISQUE$\uparrow$ & SER-FIQ$\downarrow$ & FDFR$\uparrow$ & ISM$\downarrow$ \\
\midrule
\multirow{7}{*}{VGGFace2}
 & No Def.    & 14.47 & 0.63 & 0.13 & 0.48 & 22.74 & 0.64 & 0.32 & 0.28 \\
 & Anti-DB       & 43.57 & \textbf{0.09} & \textbf{0.88} & 0.19 & 36.91 & \textbf{0.31} & \underline{0.40} & \underline{0.12} \\
 & Mist          & 34.72 & 0.32 & 0.54 & \textbf{0.13} & 30.52 & \underline{0.38} & 0.37 & \textbf{0.10} \\
 & SDS(-)        & \textbf{50.62} & 0.73 & 0.08 & 0.57 & \underline{45.38} & 0.52 & 0.23 & 0.20 \\
 & MYOPIA        & \underline{50.18} & 0.71 & 0.20 & 0.51 & \textbf{47.51} & 0.51 & 0.21 & 0.20 \\
 & VCPro         & 32.24 & 0.60 & 0.16 & 0.41 & 26.53 & 0.63 & 0.34 & 0.25 \\
 & \textbf{\method} & 44.46 & \underline{0.31} & \underline{0.79} & \underline{0.15} & 30.15 & \textbf{0.31} & \textbf{0.44} & \textbf{0.10} \\
\midrule
\multirow{7}{*}{CelebA-HQ}
 & No Def.    & 15.89 & 0.66 & 0.19 & 0.47 & 21.49 & 0.63 & 0.42 & 0.22 \\
 & Anti-DB       & \underline{48.16} & \textbf{0.05} & \textbf{0.96} & \underline{0.11} & 38.02 & \textbf{0.27} & 0.51 & \underline{0.07} \\
 & Mist          & 34.67 & 0.38 & 0.47 & 0.13 & 34.25 & 0.33 & \textbf{0.56} & \textbf{0.06} \\
 & SDS(-)        & 47.68 & 0.75 & 0.14 & 0.55 & \underline{40.13} & 0.37 & \underline{0.52} & 0.09 \\
 & MYOPIA        & \textbf{52.54} & 0.76 & 0.17 & 0.52 & \textbf{49.34} & 0.40 & 0.48 & 0.12 \\
 & VCPro         & 42.94 & 0.62 & 0.29 & 0.36 & 24.26 & 0.62 & 0.51 & 0.18 \\
 & \textbf{\method} & 45.88 & \underline{0.27} & \underline{0.84} & \textbf{0.09} & 27.29 & \underline{0.28} & 0.48 & \textbf{0.06} \\
\bottomrule
\end{tabular}
\caption{Protection effectiveness under additional prompts on VGGFace2 and CelebA-HQ.
P3: \textit{a photo of sks person looking at the mirror}; P4: \textit{a photo of sks person in front of eiffel tower}; P5: \textit{a close-up photo of sks person, high details}; P6: \textit{a photo of sks person stand in front of the table holding a knife}.}
\label{tab:prompt3456}
\end{table*}

